# An optimization algorithm for multimodal functions inspired by collective animal behavior

Erik Cuevas[1] and Mauricio González

Departamento de Ciencias Computacionales
Universidad de Guadalajara, CUCEI
Av. Revolución 1500, Guadalajara, Jal, México
{[1]erik.cuevas, mauricio.gonzalez}@cucei.udg.mx

## Abstract

Interest in multimodal function optimization is expanding rapidly since real-world optimization problems often demand locating multiple optima within a search space. This article presents a new multimodal optimization algorithm named as the Collective Animal Behavior (CAB). Animal groups, such as schools of fish, flocks of birds, swarms of locusts and herds of wildebeest, exhibit a variety of behaviors including swarming about a food source, milling around a central location or migrating over large distances in aligned groups. These collective behaviors are often advantageous to groups, allowing them to increase their harvesting efficiency to follow better migration routes, to improve their aerodynamic and to avoid predation. In the proposed algorithm, searcher agents are a group of animals which interact to each other based on the biological laws of collective motion. Experimental results demonstrate that the proposed algorithm is capable of finding global and local optima of benchmark multimodal optimization problems with a higher efficiency in comparison to other methods reported in the literature.

*Keywords:* Metaheuristic algorithms, Multimodal optimization, Evolutionary algorithms, Bio-inspired algorithms.

## 1. Introduction

A large number of real-world problems can be considered as multimodal function optimization subjects. An objective function may have several global optima, i.e. several points holding objective function values which are equal to the global optimum. Moreover, it may exhibit some other local optima points whose objective function values lay nearby a global optimum. Since the mathematical formulation of a real-world problem often produces a multimodal optimization issue, finding all global or even these local optima would provide to the decision makers multiple options to choose from [1].

Several methods have recently been proposed for solving the multimodal optimization problem. They can be divided into two main categories: deterministic and stochastic (metaheuristic) methods. When facing complex multimodal optimization problems, deterministic methods, such as gradient descent method, the quasi-Newton method and the Nelder-Mead's simplex method, may get easily trapped into the local optimum as a result of deficiently exploiting local information. They strongly depend on a priori information about the objective function, yielding few reliable results.

Metaheuristic algorithms have been developed combined rules and randomness mimicking several phenomena. These phenomena include evolutionary processes (e.g., the evolutionary algorithm proposed by Fogel et al. [2], De Jong [3], and Koza [4] and the genetic algorithms (GAs) proposed by Holland [5] and Goldberg [6]), immunological systems (e.g., the artificial immune systems proposed by de Castro et al. [7]), physical processes (e.g., simulated annealing proposed by Kirkpatrick et al. [8], electromagnetism-like proposed by İlker et al. [9] and the gravitational search algorithm proposed by Rashedi et al. [10] ) and the musical process of searching for a perfect state of harmony (proposed by Geem et al. [11], Lee and Geem [12], Geem [13] and Gao et al. [14]).

Traditional GA's perform well for locating a single optimum but fail to provide multiple solutions. Several methods have been introduced into the GA's scheme to achieve multimodal function optimization, such as sequential fitness sharing [15,16], deterministic crowding [17], probabilistic crowding [18], clustering based niching [19], clearing procedure [20], species conserving genetic algorithm [21], and elitist-population strategies [22]. However, algorithms based on the GA's do not guarantee convergence to global optima because of their poor exploitation capability. GA's exhibit other drawbacks such as the premature convergence which results from the loss of diversity in the population and becomes a common problem when the search continues for several generations. Such drawbacks [23] prevent the GA's from practical interest for several applications.

---
[1] Corresponding author, Tel +52 33 1378 5900,  ext. 27714, E-mail: erik.cuevas@cucei.udg.mx

Using a different metaphor, other researchers have employed Artificial Immune Systems (AIS) to solve the multimodal optimization problems. Some examples are the clonal selection algorithm [24] and the artificial immune network (AiNet) [25, 26]. Both approaches use some operators and structures which attempt to algorithmically mimic the natural immune system's behavior of human beings and animals.

On other hand, many studies have been inspired by animal behavior phenomena in order to develop optimization techniques such as the Particle swarm optimization (PSO) algorithm which models the social behavior of bird flocking or fish schooling [27]. In recent years, there have been several attempts to apply the PSO to multi-modal function optimization problems [28,29]. However, the performance of such approaches presents several flaws when it is compared to the other multi-modal metaheuristic counterparts [26].

Recently, the concept of individual-organization [30, 31] has been widely used to understand collective behavior of animals. The central principle of individual-organization is that simple repeated interactions between individuals can produce complex behavioral patterns at group level [30, 32, 33]. Such inspiration comes from behavioral patterns seen in several animal groups, such as ant pheromone trail networks, aggregation of cockroaches and the migration of fish schools, which can be accurately described in terms of individuals following simple sets of rules [34]. Some examples of these rules [33, 35] include keeping current position (or location) for best individuals, local attraction or repulsion, random movements and competition for the space inside of a determined distance. On the other hand, new studies have also shown the existence of collective memory in animal groups [36-38]. The presence of such memory establishes that the previous history, of group structure, influences the collective behavior exhibited in future stages. Therefore, according to these new developments, it is possible to model complex collective behaviors by using simple individual rules and configuring a general memory.

This paper proposes a new optimization algorithm inspired by the collective animal behavior. In this algorithm, the searcher agents are a group of animals that interact to each other based on simple behavioral rules which are modeled as mathematical operators. Such operations are applied to each agent considering that the complete group has a memory which stores its own best positions seen so far by applying a competition principle. The proposed approach has also been compared to some other well-known metaheuristic search methods. The obtained results confirm a high performance of the proposed method for solving various benchmark functions.

This paper is organized as follows: Section 2 introduces the basic biologic aspects of the algorithm. In Section 3, the proposed algorithm and its characteristics are described. A comparative study is presented in Section 4 and finally in Section 5 the conclusions are discussed.

## 2. Biologic fundaments

The remarkable collective behavior of organisms such as swarming ants, schooling fish and flocking birds has long captivated the attention of naturalists and scientists. Despite a long history of scientific investigation, just recently we are beginning to decipher the relationship between individuals and group-level properties [39]. Grouping individuals often have to make rapid decisions about where to move or what behavior to perform, in uncertain and dangerous environments. However, each individual typically has only relatively local sensing ability [40]. Groups are, therefore, often composed of individuals that differ with respect to their informational status and individuals are usually not aware of the informational state of others [41], such as whether they are knowledgeable about a pertinent resource, or of a threat.

Animal groups are based on a hierarchic structure [42] which differentiates individuals according to a fitness principle known as Dominance [43]. Such concept represents the domain of some individuals within a group and occurs when competition for resources leads to confrontation. Several studies [44,45] have found that such animal behavior lead to stable groups with better cohesion properties among individuals.

Recent studies have illustrated how repeated interactions among grouping animals scale to collective behavior. They have also remarkably revealed, that collective decision-making mechanisms across a wide range of animal group types, ranging from insects to birds (and even among humans in certain circumstances) seem to share similar functional characteristics [30,34,46]. Furthermore, at a certain level of description, collective decision-making in organisms shares essential common features such as a general memory. Although some differences may arise, there are good reasons to increase communication between researchers working in collective animal behavior and those involved in cognitive science [33].

Despite the variety of behaviors and motions of animal groups, it is possible that many of the different collective behavioral patterns are generated by simple rules followed by individual group members. Some authors have developed different models, such as the self-propelled particle (SPP) model which attempts to capture the collective behavior of animal groups in terms of interactions between group members following a diffusion process [47-50].

On other hand, following a biological approach, Couzin et al. [33,34] have proposed a model in which individual animals follow simple rules of thumb: (1) keep the position of best individuals; (2) move from or to nearby neighbors (local attraction or repulsion); (3) move randomly and (4) compete for the space inside of a determined distance. Each individual thus admits three different movements: attraction, repulsion or random, while holds two kinds of states: preserve the position or compete for a determined position. In the model, the movement experimented by each individual is decided randomly (according to an internal motivation), meanwhile the states are assumed according to a fixed criteria.

The dynamical spatial structure of an animal group can be explained in terms of its history [47]. Despite this, the majority of the studies have failed in considering the existence of memory in behavioral models. However, recent researches [36,51] have also shown the existence of collective memory in animal groups. The presence of such memory establishes that the previous history of the group structure, influences the collective behavior exhibited in future stages. Such memory can contain the position of special group members (the dominant individuals) or the averaged movements produced by the group.

According to these new developments, it is possible to model complex collective behaviors by using simple individual rules and setting a general memory. In this work, the behavioral model of animal groups is employed for defining the evolutionary operators through the proposed metaheuristic algorithm. A memory is incorporated to store best animal positions (best solutions) considering a competition-dominance mechanism.

## 3. Collective Animal Behaviour Algorithm (CAB)

The CAB algorithm assumes the existence of a set of operations that resembles the interaction rules that model the collective animal behavior. In the approach, each solution within the search space represents an animal position. The "fitness value" refers to the animal dominance with respect to the group. The complete process mimics the collective animal behavior.

The approach in this paper implements a memory for storing best solutions (animal positions) mimicking the aforementioned biologic process. Such memory is divided into two different elements, one for maintaining the best found positions in each generation ($\mathbf{M}_g$) and the other for storing best history positions during the complete evolutionary process ($\mathbf{M}_h$).

### 3.1 Description of the CAB algorithm

Likewise other metaheuristic approaches, the CAB algorithm is also an iterative process. It starts by initializing the population randomly, i.e. generating random solutions or animal positions. The following four operations are thus applied until the termination criterion is met, i.e. the iteration number *NI* is reached as follows:

1. Keep the position of the best individuals.
2. Move from or nearby neighbors (local attraction and repulsion).
3. Move randomly.
4. Compete for the space inside of a determined distance (updating the memory).

### 3.1.1 Initializing the population

The algorithm begins by initializing a set $\mathbf{A}$ of $N_p$ animal positions ( $\mathbf{A} = \{\mathbf{a}_1, \mathbf{a}_2, \ldots, \mathbf{a}_{N_p}\}$ ). Each animal position $\mathbf{a}_i$ is a $D$-dimensional vector containing the parameter values to be optimized, which are randomly and uniformly distributed between the pre-specified lower initial parameter bound $a_j^{low}$ and the upper initial parameter bound $a_j^{high}$.

$$a_{j,i} = a_j^{low} + \text{rand}(0,1) \cdot (a_j^{high} - a_j^{low});$$
$$j = 1, 2, \ldots, D; \quad i = 1, 2, \ldots, N_p. \tag{1}$$

with $j$ and $i$ being the parameter and individual indexes respectively. Hence, $a_{j,i}$ is the $j$th parameter of the $i$th individual.

All the initial positions $\mathbf{A}$ are sorted according to the fitness function (dominance) to form a new individual set $\mathbf{X} = \{\mathbf{x}_1, \mathbf{x}_2, \ldots, \mathbf{x}_{N_p}\}$, so that we can choose the best $B$ positions and store them in the memory $\mathbf{M}_g$ and $\mathbf{M}_h$. The fact that both memories share the same information is only allowed at this initial stage.

### 3.1.2 Keep the position of the best individuals.

Analogously to the biological metaphor, this behavioral rule, typical in animal groups, is implemented as an evolutionary operation in our approach. In this operation, the first $B$ elements of the new animal position set $\mathbf{A}$ ( $\{\mathbf{a}_1, \mathbf{a}_2, \ldots, \mathbf{a}_B\}$ ) are generated. Such positions are computed by the values contained in the historic memory $\mathbf{M}_h$ considering a slight random perturbation around them. This operation can be modelled as follows:

$$\mathbf{a}_l = \mathbf{m}_h^l + \mathbf{v} \tag{2}$$

where $l \in \{1, 2, \ldots, B\}$ while $\mathbf{m}_h^l$ represents the $l$-element of the historic memory $\mathbf{M}_h$ and $\mathbf{v}$ is a random vector holding an appropriate small length.

### 3.1.3 Move from or to nearby neighbours.

From the biological inspiration, where animals experiment a random local attraction or repulsion according to an internal motivation, we implement the evolutionary operators that mimic them. For this operation, a uniform random number $r_m$ is generated within the range [0,1]. If $r_m$ is less than a threshold $H$, a determined individual position is moved (attracted or repelled) considering the nearest best historical value of the group (the nearest position contained in $\mathbf{M}_h$) otherwise it is considered the nearest best value in the group of the current generation (the nearest position contained in $\mathbf{M}_g$). Therefore such operation can be modeled as follows:

$$\mathbf{a}_i = \begin{cases} \mathbf{x}_i \pm r \cdot (\mathbf{m}_h^{nearest} - \mathbf{x}_i) & \text{with probability } H \\ \mathbf{x}_i \pm r \cdot (\mathbf{m}_g^{nearest} - \mathbf{x}_i) & \text{with probability } (1\text{-}H) \end{cases} \tag{3}$$

where $i \in \{B+1, B+2, \ldots, N_p\}$, $\mathbf{m}_h^{nearest}$ and $\mathbf{m}_g^{nearest}$ represent the nearest elements of $\mathbf{M}_h$ and $\mathbf{M}_g$ to $\mathbf{x}_i$, while $r$ is a random number between [-1,1]. Therefore, if $r>0$, the individual position $\mathbf{x}_i$ is attracted to the position $\mathbf{m}_h^{nearest}$ or $\mathbf{m}_g^{nearest}$, otherwise such movement is considered as a repulsion.

### 3.1.4 Move randomly.

Following the biological model, under some probability $P$ an animal randomly changes its position. Such behavioral rule is implemented considering the next expression:

$$\mathbf{a}_i = \begin{cases} \mathbf{r} & \text{with probability } P \\ \mathbf{x}_i & \text{with probability } (1\text{-}P) \end{cases} \tag{4}$$

being $i \in \{B+1, B+2, \ldots, N_p\}$ and $\mathbf{r}$ a random vector defined within the search space. This operator is similar to re-initialize the particle in a random position as it is done by Eq. (1).

### 3.1.5. Compete for the space inside of a determined distance (updating the memory).

Once the operations to preserve the position of the best individuals, to move from or to nearby neighbors and to move randomly, have all been applied to the all $N_p$ animal positions, generating $N_p$ new positions, it is necessary to update the memory $\mathbf{M}_h$.

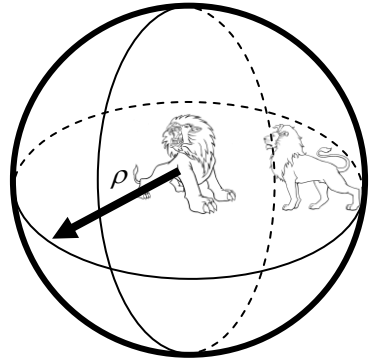

**Fig. 1.** Dominance concept, presented when two animals confront each other inside of a $\rho$ distance

In order to update de memory $\mathbf{M}_h$, the concept of dominance is used. Animals that interact in a group keep a minimum distance among them. Such distance $\rho$ depends on how aggressive the animal behaves [43,51]. Hence, when two animals confront each other inside of such distance, the most dominant individual prevails as the other withdraws. Figure 1 shows this process.

In the proposed algorithm, the historic memory $\mathbf{M}_h$ is updated considering the following procedure:

1. The elements of $\mathbf{M}_h$ and $\mathbf{M}_g$ are merged into $\mathbf{M}_U$ ( $\mathbf{M}_U = \mathbf{M}_h \cup \mathbf{M}_g$ ).
2. Each element $\mathbf{m}_U^i$ of the memory $\mathbf{M}_U$, it is compared pair-wise with the remainder memory elements ( $\{\mathbf{m}_U^1, \mathbf{m}_U^2, \ldots, \mathbf{m}_U^{2B-1}\}$ ). If the distance between both elements is less than $\rho$, the element holding a better performance in the fitness function will prevail meanwhile the other will be removed.
3. From the resulting elements of $\mathbf{M}_U$ (as they are obtained in step 2), the $B$ best value is selected to integrate the new $\mathbf{M}_h$.

Unsuitable values of $\rho$ result in a lower convergence rate, longer computation time, larger function evaluation number, convergence to a local maximum or unreliability of solutions. The $\rho$ value is computed considering the following equation:

$$\rho = \frac{\prod_{j=1}^{D}(a_j^{high} - a_j^{low})}{10 \cdot D} \tag{5}$$

where $a_j^{low}$ and $a_j^{high}$ represent the pre-specified lower bound and the upper bound of the *j*-parameter respectively, within an *D*-dimensional space.

### 3.1.6. Computational procedure

The computational procedure for the proposed algorithm can be summarized as follows:

Step 1: Set the parameters $N_p$ , *B*, *H*, *P* and *NI*.

Step 2: Generate randomly the position set $\mathbf{A} = \{\mathbf{a}_1, \mathbf{a}_2, \ldots, \mathbf{a}_{N_p}\}$ using Eq.1

Step 3: Sort **A**, according to the objective function (dominance), building $\mathbf{X} = \{\mathbf{x}_1, \mathbf{x}_2, \ldots, \mathbf{x}_{N_p}\}$ .

Step 4: Choose the first *B* positions of **X** and store them into the memory $\mathbf{M}_g$ .

Step 5: Update $\mathbf{M}_h$ according to section 3.1.5 (for the first iteration $\mathbf{M}_h = \mathbf{M}_g$ ).

Step 6: Generate the first *B* positions of the new solution set $\mathbf{A} = \{\mathbf{a}_1, \mathbf{a}_2, \ldots, \mathbf{a}_B\}$ . Such positions correspond to elements of $\mathbf{M}_h$ making a slight random perturbation around them.

$\mathbf{a}_i = \mathbf{m}_h^i + \mathbf{v}$ ; being **v** a random vector holding an appropriate small length.

Step 7: Generate the rest of the **A** elements using the attraction, repulsion and random movements.

for *i*=*B*+1: $N_p$

    if ( $r_1 < 1$-*P*) then

    *attraction and repulsion movement*

      { if ( $r_2 < H$) then

        $\mathbf{a}_i = \mathbf{x}_i \pm r \cdot (\mathbf{m}_h^{nearest} - \mathbf{x}_i)$

       else if

        $\mathbf{a}_i = \mathbf{x}_i \pm r \cdot (\mathbf{m}_g^{nearest} - \mathbf{x}_i)$

      }

    else if

    *random movement*

      {

       $\mathbf{a}_i = \mathbf{r}$

      }

    end for

where $r_1, r_2, r \in \text{rand}(0,1)$ .

Step 8: If *NI* is completed, the process is thus completed; otherwise go back to step 3.

### 3.1.7. Optima determination

Just after the optimization process has finished, an analysis of the final $\mathbf{M}_h$ memory is executed in order to find the global and significant local minima. For it, a threshold value $T_h$ is defined to decide which elements will be considered as a significant local minimum. Such threshold is thus computed as:

           

$$T_h = \frac{\max_{fitness}(\mathbf{M_h})}{6} \qquad (6)$$

where $\max_{fitness}(\mathbf{M_h})$ represents the best fitness value among $\mathbf{M}_h$ elements. Therefore, memory elements whose fitness values are greater than $T_h$ will be considered as global and local optima as other elements are discarded.

### 3.1.7. Numerical example

In order to demonstrate the algorithm's step-by-step operation, a numerical example has been set by applying the proposed method to optimize a simple function which is defined as follows:

$$f(x_1, x_2) = e^{-(x_1-4)^2 - (x_2-4)^2} + e^{-(x_1+4)^2 - (x_2-4)^2} + 2 \cdot e^{-(x_1)^2 + (x_2)^2} + 2 \cdot e^{-(x_1)^2 - (x_2+4)^2} \qquad (7)$$

Considering the interval of $-5 \le x_1, x_2 \le 5$, the function possesses two global maxima of value 2 at $(x_1, x_2) = (0,0)$ and $(0,-4)$. Likewise, it holds two local minima of value 1 at $(-4,4)$ and $(4,4)$. Fig. 2a shows the 3D plot of this function. The parameters for the CAB algorithm are set as: $N_p = 10$, $B$=4, $H$=0.8, $P$=0.1, $\rho = 3$ and $NI$=30.

Like all evolutionary approaches, CAB is a population-based optimizer that attacks the starting point problem by sampling the objective function at multiple, randomly chosen, initial points. Therefore, after setting parameter bounds that define the problem domain, 10 ( $N_p$ ) individuals $(\mathbf{i}_1, \mathbf{i}_2, ..., \mathbf{i}_{10})$ are generated using Eq. 1. Following an evaluation of each individual through the objective function (Eq. 5), all are sorted decreasingly in order to build vector $\mathbf{X} = (\mathbf{x}_1, \mathbf{x}_2, ..., \mathbf{x}_{10})$. Fig. 2b depicts the initial individual distribution in the search space. Then, both memories $\mathbf{M}_g$ $(\mathbf{m}_g^1, ..., \mathbf{m}_g^4)$ and $\mathbf{M}_h$ $(\mathbf{m}_h^1, ..., \mathbf{m}_h^4)$ are filled with the first four ($B$) elements present in $\mathbf{X}$. Such memory elements are represented by solid points in Fig 2c.

The new 10 individuals $(\mathbf{a}_1, \mathbf{a}_2, ..., \mathbf{a}_{10})$ are evolved at each iteration following three different steps: 1. Keep the position of best individuals. 2. Move from or nearby neighbors and 3. Move randomly. The first new four elements $(\mathbf{a}_1, \mathbf{a}_2, \mathbf{a}_3, \mathbf{a}_4)$ are generated considering the first step (Keeping the position of best individuals). Following such step, new individual positions are calculated as perturbed versions of all the elements which are contained in the $\mathbf{M}_h$ memory (that represent the best individuals known so far). Such perturbation is done by using $\mathbf{a}_l = \mathbf{m}_h^l + \mathbf{v}$ $(l \in 1, ..., 4)$. Fig. 2d shows a comparative view between the memory element positions and the perturbed values of $(\mathbf{a}_1, \mathbf{a}_2, \mathbf{a}_3, \mathbf{a}_4)$.

The remaining 6 new positions $(\mathbf{a}_5, ..., \mathbf{a}_{10})$ are individually computed according to step 2 and 3. For such operation, a uniform random number $r_1$ is generated within the range [0, 1]. If $r_1$ is less than 1-$P$, the new position $\mathbf{a}_j$ $(j \in 5, ..., 10)$ is generated through step 2; otherwise, $\mathbf{a}_j$ is obtained from a random re-initialization (step 3) between search bounds.

In order to calculate a new position $\mathbf{a}_j$ at step 2, a decision must be made on whether it should be generated by using the elements of $\mathbf{M}_h$ or $\mathbf{M}_g$. For such decision, a uniform random number $r_2$ is generated within the range [0, 1]. If $r_2$ is less than $H$, the new position $\mathbf{a}_j$ is generated by using $\mathbf{x}_j \pm r \cdot (\mathbf{m}_h^{nearest} - \mathbf{x}_j)$; otherwise, $\mathbf{a}_j$ is obtained by considering $\mathbf{x}_j \pm r \cdot (\mathbf{m}_g^{nearest} - \mathbf{x}_j)$. Where $\mathbf{m}_h^{nearest}$ and $\mathbf{m}_g^{nearest}$ represent the closest elements to $\mathbf{x}_j$ in memory $\mathbf{M}_h$ and $\mathbf{M}_g$ respectively. In the first iteration, since there is not available information from previous steps, both memories $\mathbf{M}_h$ and $\mathbf{M}_g$ share the same information which is only allowed at this initial stage. Fig. 2e shows graphically the whole procedure employed by step 2 in order to calculate the new individual position $\mathbf{a}_8$ whereas Fig. 2f presents the positions of all new individuals $(\mathbf{a}_1, \mathbf{a}_2, \ldots, \mathbf{a}_{10})$.

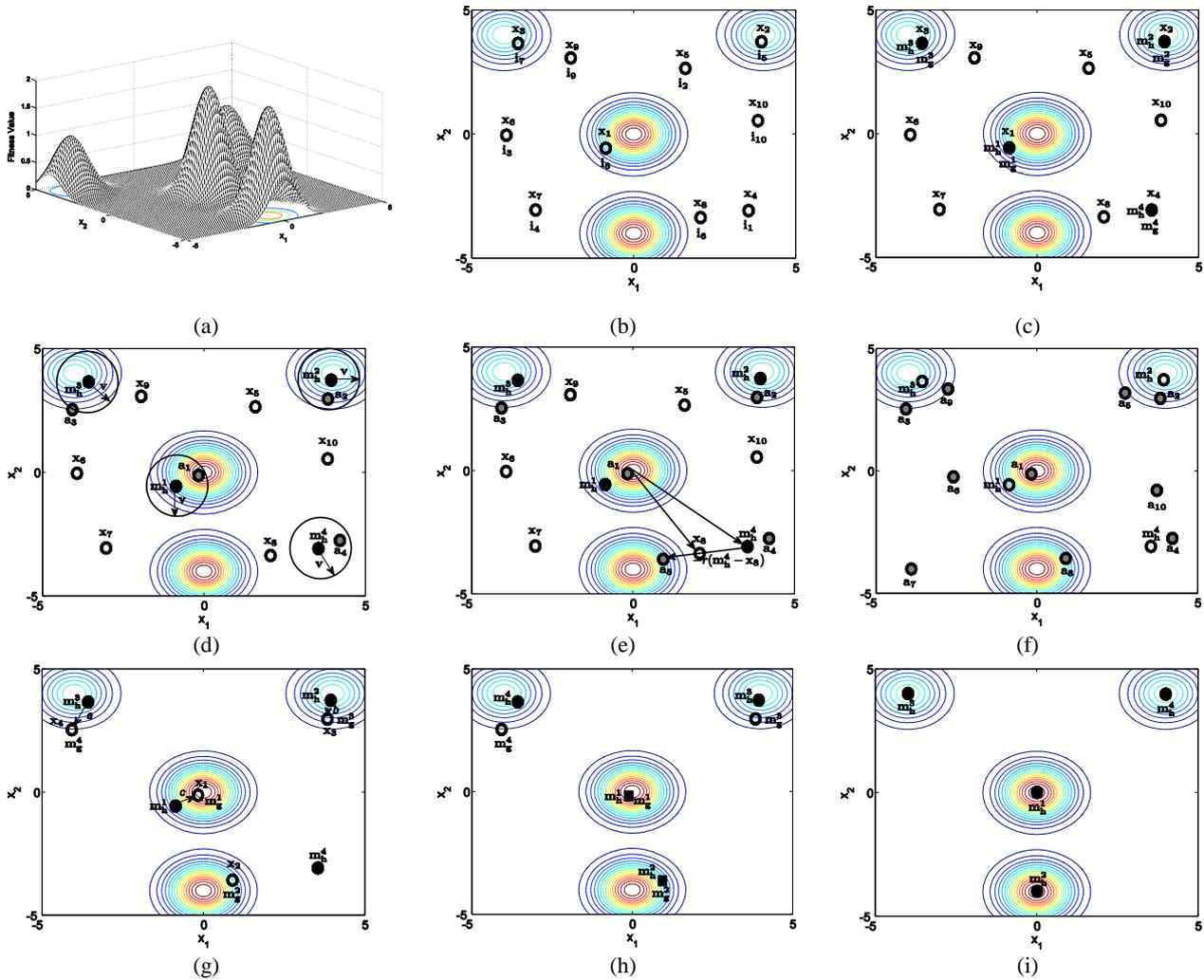

**Fig. 2.** CAB numerical example: (a) 3D plot of the function used as example. (b) Initial individual distribution. (c) Initial configuration of memories $\mathbf{M}_g$ and $\mathbf{M}_h$. (d) The computation of the first four individuals $(\mathbf{a}_1, \mathbf{a}_2, \mathbf{a}_3, \mathbf{a}_4)$. (e) It shows the procedure employed by step 2 in order to calculate the new individual position $\mathbf{a}_8$. (f) Positions of all new individuals $(\mathbf{a}_1, \mathbf{a}_2, \ldots, \mathbf{a}_{10})$. (g) Application of the dominance concept over elements of $\mathbf{M}_g$ and $\mathbf{M}_h$. (h) Final memory configurations of $\mathbf{M}_g$ and $\mathbf{M}_h$ after the first iteration. (i) Final memory configuration of $\mathbf{M}_h$ after 30 iterations.

Finally, after all new positions $(\mathbf{a}_1, \mathbf{a}_2, \ldots, \mathbf{a}_{10})$ have been calculated, memories $\mathbf{M}_h$ and $\mathbf{M}_g$ must be updated. In order to update $\mathbf{M}_h$, new calculated positions $(\mathbf{a}_1, \mathbf{a}_2, \ldots, \mathbf{a}_{10})$ are arranged according to their fitness values by building vector $\mathbf{X} = (\mathbf{x}_1, \mathbf{x}_2, \ldots, \mathbf{x}_{10})$. Then, the elements of $\mathbf{M}_h$ are replaced by the first four elements in $\mathbf{X}$ (the best individuals of its generation). In order to calculate the new elements of $\mathbf{M}_h$, current elements of $\mathbf{M}_h$ (the present values) and $\mathbf{M}_g$ (the updated values) are merged into $\mathbf{M}_U$. Then, by using the dominance concept (explained in section 3.1.5) over $\mathbf{M}_U$, the best four values are selected to replace the elements in $\mathbf{M}_g$. Figure 2g and 2h show the updating procedure for both memories. Applying the dominance (see Fig 2g), since the distances $a = dist(\mathbf{m}_h^3, \mathbf{m}_g^4)$, $b = dist(\mathbf{m}_h^2, \mathbf{m}_g^3)$ and $c = dist(\mathbf{m}_h^1, \mathbf{m}_g^1)$ are less than $\rho = 3$, elements with better fitness evaluation will build the new memory $\mathbf{M}_h$. Fig. 2h depicts final memory configurations. The circles and solid circles points represent the elements of $\mathbf{M}_g$ and $\mathbf{M}_h$ respectively whereas the bold squares perform as elements shared by both memories. Therefore, if the complete procedure is repeated over 30 iterations, the memory $\mathbf{M}_h$ will contain the 4 global and local maxima as elements. Fig. 2i depicts the final configuration after 30 iterations.

## 4. Experimental results

In this section, the performance of the proposed algorithm is tested. Section 4.1 describes the experiment methodology. Sections 4.2, and 4.3 report on a comparison between the CAB experimental results and other multimodal metaheuristic algorithms for different kinds of optimization problems.

| Function | Search space | Sketch |
|---|---|---|
| $f_1 = \sin^6(5\pi x)$ <br><br> **Deb's function** <br> **5 optima** | $x \in [0,1]$ |  |
| $f_2(x) = 2^{-2((x-0.1)/0.9)^2} \cdot \sin(5\pi x)$ <br><br> **Deb's decreasing function** <br> **5 optima** | $x \in [0,1]$ |  |
| $f_3(z) = \dfrac{1}{1+\lvert z^6 +1\rvert}$ <br><br> **Roots function** <br> **6 optima** | $z \in C,\ z = x_1 + ix_2$ <br> $x_1, x_2 \in [-2,2]$ |  |
| $f_4(x_1,x_2) = x_1 \sin(4\pi x_1) - x_2 \sin(4\pi x_2 + \pi) + 1$ <br><br> **Two dimensional multi-modal function** <br> **100 optima** | $x_1, x_2 \in [-2,2]$ |  |

**Table 1.** The test suite of multimodal functions for Experiment 4.2

### 4.1 Experiment methodology

In this section, we will examine the search performance of the proposed CAB by using a test suite of 8 benchmark functions with different complexities. They are listed in Tables 1 and 2. The suite mainly contains some representative, complicated and multimodal functions with several local optima. These functions are normally regarded as difficult to be optimized as they are particularly challenging to the applicability and efficiency of multimodal metaheuristic algorithms. The performance measurements considered at each experiment are the following:

- The consistency of locating all known optima; and

- The averaged number of objective function evaluations that are required to find such optima (or the running time under the same condition).

| Function | Search space | Sketch |
|---|---|---|
| $f_5(x_1,x_2) = -(20 + x_1^2 + x_2^2 - 10(\cos(2\pi x_1) + \cos(2\pi x_2)))$  **Rastringin's function 100 optima** | $x_1, x_2 \in [-10,10]$ | 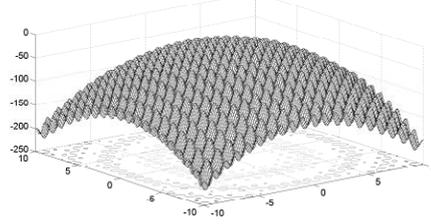 |
| $f_6(x_1,x_2) = -\prod_{i=1}^{2}\sum_{j=1}^{5}\cos((j+1)x_i + j)$  **Shubert function 18 optima** | $x_1, x_2 \in [-10,10]$ | 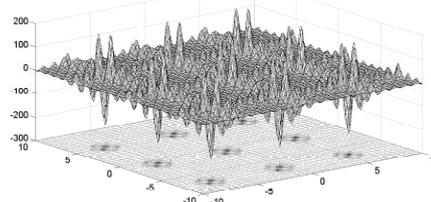 |
| $f_7(x_1,x_2) = \frac{1}{4000}\sum_{i=1}^{2}x_i^2 - \prod_{i=1}^{2}\cos\left(\frac{x_i}{\sqrt{2}}\right) + 1$  **Griewank function 100 optima** | $x_1, x_2 \in [-100,100]$ | 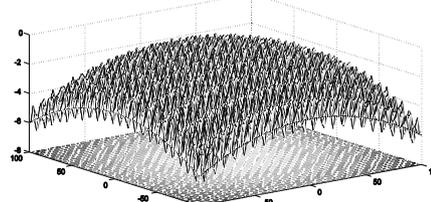 |
| $f_8(x_1,x_2) = \frac{\cos(0.5x_1) + \cos(0.5x_2)}{4000} + \cos(10x_1)\cos(10x_2)$  **Modified Griewank function 100 optima** | $x_1, x_2 \in [0,120]$ | 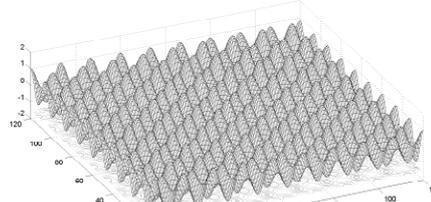 |

**Table 2.** The test suite of multimodal functions used in the Experiment 4.3.

The experiments compare the performance of CAB against the Deterministic Crowding [17], the Probabilistic Crowding [18], the Sequential Fitness Sharing [15], the Clearing Procedure [20], the Clustering Based Niching (CBN) [19], the Species Conserving Genetic Algorithm (SCGA) [21], the Elitist-population strategy (AEGA) [22], the Clonal Selection algorithm [24] and the artificial immune network (AiNet) [25].

Since the approach solves real-valued multimodal functions, we have used, in the GA-approaches, consistent real coding variable representation, uniform crossover and mutation operators for each algorithm seeking a fair comparison. The crossover probability Pc=0.8 and the mutation probability Pm=0.1 have been used. We use the standard tournament selection operator with a tournament size=2 in our implementation of Sequential Fitness Sharing, Clearing Procedure, CBN, Clonal Selection algorithm, and SCGA. On the other hand, the parameter values for the aiNet algorithm have been defined as suggested in [25], with the mutation strength $\beta = 100$, the suppression threshold $\sigma_{s(aiNet)} = 0.2$ and the update rate $d = 40\%$.

In the case of the CAB algorithm, the parameters are set to $N_p = 200$, $B$=100, $P$=0.8 and $H$=0.6. Once they have been all experimentally determined, they are kept for all the test functions through all experiments.

To avoid relating the optimization results to the choice of a particular initial population and to conduct fair comparisons, we perform each test 50 times, starting from various randomly selected points in the search domain as it is commonly given in the literature. An optimum $o_j$ is considered as found if $\exists\, x_i \in Pop(k=T) \big| d(x_i, o_j) < 0.005$, where $Pop(k=T)$ is the complete population at the end of the run $T$ and $x_i$ is an individual in $Pop(k=T)$.

All algorithms have been tested in MatLAB© over the same Dell Optiplex GX260 computer with a Pentium-4 2.66G-HZ processor, running Windows XP operating system over 1Gb of memory. Next sections present experimental results for multimodal optimization problems which have been divided into two groups with different purposes. The first one consists of functions with smooth landscapes and well defined optima (local and global values), while the second gathers functions holding rough landscapes and complex location optima.

### 4.2. Comparing CAB performance for smooth landscapes functions

This section presents a performance comparison for different algorithms solving multimodal problems $f_1 - f_4$ in Table 1. The aim is to determine whether CAB is more efficient and effective than other existing algorithms for finding all multiple optima of $f_1 - f_4$. The stopping criterion analyzes if the number identified optima cannot be further increased over 10 successive generations after the first 100 generations, then the execution will be stopped. Four measurements have been employed to evaluate the performance:

• The average of optima found within the final population (NO);
• The average distance between multiple optima detected by the algorithm and their closest individuals in the final population (DO);
• The average of function evaluations (FE); and
• The average of execution time in seconds (ET).

Table 3 provides a summarized performance comparison among several algorithms. Best results have been bold-faced. From the NO measure, CAB always finds better or equally optimal solutions for the multimodal problems $f_1 - f_4$. It is evident that each algorithm can find all optima of $f_1$. For function $f_2$, only AEGA, Clonal Selection algorithm, aiNet, and CAB can eventually find all optima each time. For function $f_3$, Clearing Procedure, SCGA, AEGA and CAB can get all optima at each run. For function $f_4$, Deterministic Crowding leads to premature convergence and all other algorithms cannot get any better results but CAB yet can find all multiple optima 48 times in 50 runs and its average successful rate for each run is higher than 99%. By analyzing the DO measure in Table 3, CAB has obtained the best score for all the multimodal problems except for $f_3$. In the case of $f_3$, the solution precision of CAB is only worse than that of Clearing Procedure. On the other hand, CAB has smaller standard deviations in the NO and DO measures than all other algorithms and hence its solution is more stable.

From the FE measure in Table 3, it is clear that CAB needs fewer function evaluations than other algorithms considering the same termination criterion. Recall that all algorithms use the same conventional crossover and mutation operators. It can be easily deduced from results that the CAB algorithm is able to produce better search positions (better compromise between exploration and exploitation), in a more efficient and effective way than other multimodal search strategies.

To validate that CAB improvement over other algorithms as a result of CAB producing better search positions over iterations, Fig.3 shows the comparison of CAB and other multimodal algorithms for $f_4$. The initial populations for all algorithms have 200 individuals. In the final population of CAB, the 100 individuals belonging to the $\mathbf{M}_h$ memory correspond to the 100 multiple optima, while, on the contrary, the final population of the other nine algorithms fail consistently in finding all optima, despite they have superimposed several times over some previously found optima.

| function | Algorithm | NO | DO | FE | ET |
|---|---|---|---|---|---|
| $f_1$ | Deterministic crowding | **5(0)** | $1.52\times10^{-4}(1.38\times10^{-4})$ | 7,153 (358) | 0.091(0.013) |
| | Probabilistic crowding | **5(0)** | $3.63\times10^{-4}(6.45\times10^{-5})$ | 10,304(487) | 0.163(0.011) |
| | Sequential fitness sharing | **5(0)** | $4.76\times10^{-4}(6.82\times10^{-5})$ | 9,927(691) | 0.166(0.028) |
| | Clearing procedure | **5(0)** | $1.27\times10^{-2}(2.13\times10^{-5})$ | 5,860(623) | 0.128(0.021) |
| | CBN | **5(0)** | $2.94\times10^{-4}(4.21\times10^{-5})$ | 10,781(527) | 0.237(0.019) |
| | SCGA | **5(0)** | $1.16\times10^{-4}(3.11\times10^{-5})$ | 6,792(352) | 0.131(0.009) |
| | AEGA | **5(0)** | $4.6\times10^{-5}(1.35\times10^{-5})$ | 2,591(278) | 0.039(0.007) |
| | Clonal selection algorithm | **5(0)** | $1.99\times10^{-4}(8.25\times10^{-5})$ | 15,803(381) | 0.359(0.015) |
| | AiNet | **5(0)** | $1.28\times10^{-4}(3.88\times10^{-5})$ | 12,369(429) | 0.421(0.021) |
| | CAB | **5(0)** | $\mathbf{1.69\times10^{-5}(5.2\times10^{-6})}$ | **1,776(125)** | **0.020(0.009)** |
| $f_2$ | Deterministic crowding | 3.53(0.73) | $3.61\times10^{-3}(6.88\times10^{-4})$ | 6,026 (832) | 0.271(0.06) |
| | Probabilistic crowding | 4.73(0.64) | $2.82\times10^{-3}(8.52\times10^{-4})$ | 10,940(9517) | 0.392(0.07) |
| | Sequential fitness sharing | 4.77(0.57) | $2.33\times10^{-3}(4.36\times10^{-4})$ | 12,796(1,430) | 0.473(0.11) |
| | Clearing procedure | 4.73(0.58) | $4.21\times10^{-3}(1.24\times10^{-3})$ | 8,465(773) | 0.326(0.05) |
| | CBN | 4.70(0.53) | $2.19\times10^{-3}(4.53\times10^{-4})$ | 14,120(2,187) | 0.581(0.14) |
| | SCGA | 4.83(0.38) | $3.15\times10^{-3}(4.71\times10^{-4})$ | 10,548(1,382) | 0.374(0.09) |
| | AEGA | **5(0)** | $1.38\times10^{-4}(2.32\times10^{-5})$ | 3,605(426) | 0.102(0.04) |
| | Clonal selection algorithm | **5(0)** | $1.37\times10^{-3}(6.87\times10^{-4})$ | 21,922(746) | 0.728(0.06) |
| | AiNet | **5(0)** | $1.22\times10^{-3}(5.12\times10^{-4})$ | 18,251(829) | 0.664(0.08) |
| | CAB | **5(0)** | $\mathbf{4.5\times10^{-5}(8.56\times10^{-6})}$ | **2,065(92)** | **0.08(0.007)** |
| $f_3$ | Deterministic crowding | 4.23(1.17) | $7.79\times10^{-4}(4.76\times10^{-4})$ | 11,009 (1,137) | 1.07(0.13) |
| | Probabilistic crowding | 4.97(0.64) | $2.35\times10^{-3}(7.14\times10^{-4})$ | 16,391(1,204) | 1.72(0.12) |
| | Sequential fitness sharing | 4.87(0.57) | $2.56\times10^{-3}(2.58\times10^{-3})$ | 14,424(2,045) | 1.84(0.26) |
| | Clearing procedure | **6(0)** | $\mathbf{7.43\times10^{-4}(4.07\times10^{-5})}$ | 12,684(1,729) | 1.59(0.19) |
| | CBN | 4.73(1.14) | $1.85\times10^{-3}(5.42\times10^{-4})$ | 18,755(2,404) | 2.03(0.31) |
| | SCGA | **6(0)** | $3.27\times10^{-4}(7.46\times10^{-5})$ | 13,814(1,382) | 1.75(0.21) |
| | AEGA | **6(0)** | $1.21\times10^{-4}(8.63\times10^{-5})$ | 6,218(935) | 0.53(0.07) |
| | Clonal selection algorithm | 5.50(0.51) | $4.95\times10^{-3}(1.39\times10^{-3})$ | 25,953(2,918) | 2.55(0.33) |
| | AiNet | 4.8(0.33) | $3.89\times10^{-3}(4.11\times10^{-4})$ | 20,335(1,022) | 2.15(0.10) |
| | CAB | **6(0)** | $9.87\times10^{-5}(1.69\times10^{-5})$ | **4,359(75)** | **0.11(0.023)** |
| $f_4$ | Deterministic crowding | 76.3(11.4) | $4.52\times10^{-3}(4.17\times10^{-3})$ | 1,861,707(329,254) | 21.63(2.01) |
| | Probabilistic crowding | 92.8(3.46) | $3.46\times10^{-3}(9.75\times10^{-4})$ | 2,638,581(597,658) | 31.24(5.32) |
| | Sequential fitness sharing | 89.9(5.19) | $2.75\times10^{-3}(6.89\times10^{-4})$ | 2,498,257(374,804) | 28.47(3.51) |
| | Clearing procedure | 89.5(5.61) | $3.83\times10^{-3}(9.22\times10^{-4})$ | 2,257,964(742,569) | 25.31(6.24) |
| | CBN | 90.8(6.50) | $4.26\times10^{-3}(1.14\times10^{-3})$ | 2,978,385(872,050) | 35.27(8.41) |
| | SCGA | 91.4(3.04) | $3.73\times10^{-3}(2.29\times10^{-3})$ | 2,845,789(432,117) | 32.15(4.85) |
| | AEGA | 95.8(1.64) | $1.44\times10^{-4}(2.82\times10^{-5})$ | 1,202,318(784,114) | 12.17(2.29) |
| | Clonal selection algorithm | 92.1(4.63) | $4.08\times10^{-3}(8.25\times10^{-3})$ | 3,752,136(191,849) | 45.95(1.56) |
| | AiNet | 93.2(7.12) | $3.74\times10^{-3}(5.41\times10^{-4})$ | 2,745,967(328,176) | 38.18(3.77) |
| | CAB | **100(2)** | $\mathbf{2.31\times10^{-4}(5.87\times10^{-6})}$ | **697,578(57,089)** | **5.78(1.26)** |

**Table 3.** Performance comparison among the multimodal optimization algorithms for the test functions $f_1 - f_4$. The standard unit in the column ET is seconds. (For all the parameters, numbers in parentheses are the standard deviations.). Bold-cased letters represents best obtained results.

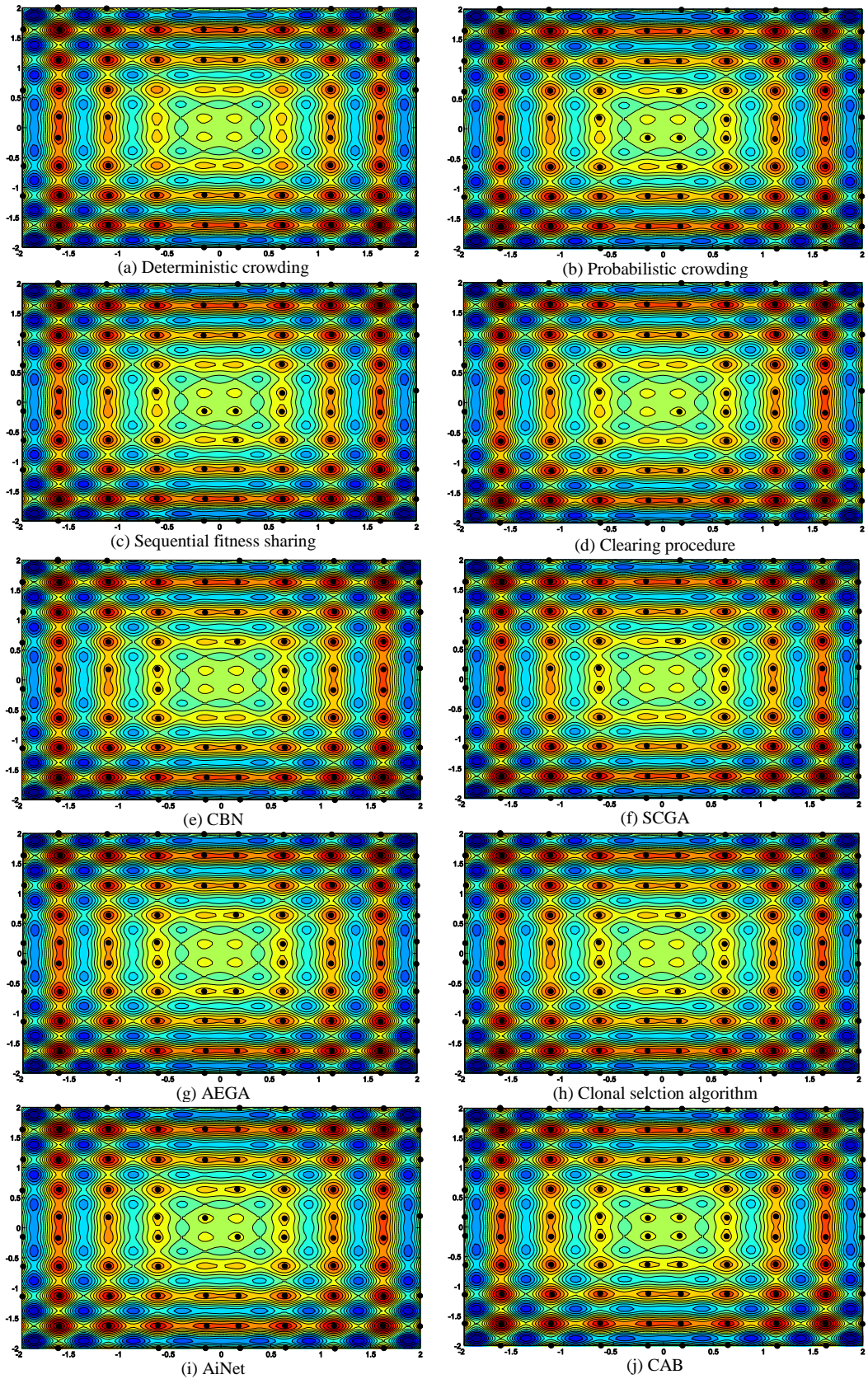

**Fig. 3.** Typical results of the maximization of $f_4$ . (a)-(j) Local and global optima located by all ten algorithms in the performance comparison.

When comparing the execution time (ET) in Table 3, CAB uses significantly less time to finish than other algorithms. The situation can be registered by the reduction of the redundancy in the $\mathbf{M}_h$ memory due to competition (dominance) criterion. All these comparisons show that CAB generally outperforms all other multimodal algorithms regarding efficacy and efficiency.

| function | Algorithm | NO | DO | FE | ET |
|---|---|---|---|---|---|
| $f_5$ | Deterministic crowding | 62.4(14.3) | $4.72\times10^{-3}(4.59\times10^{-3})$ | 1,760,199(254,341) | 14.62(2.83) |
| | Probabilistic crowding | 84.7(5.48) | $1.50\times10^{-3}(9.38\times10^{-4})$ | 2,631,627(443,522) | 34.39(5.20) |
| | Sequential fitness sharing | 76.3(7.08) | $3.51\times10^{-3}(1.66\times10^{-3})$ | 2,726,394(562,723) | 36.55(7.13) |
| | Clearing procedure | 93.6(2.31) | $2.78\times10^{-3}(1.20\times10^{-3})$ | 2,107,962(462,622) | 28.61(6.47) |
| | CBN | 87.9(7.78) | $4.33\times10^{-3}(2.82\times10^{-3})$ | 2,835,119(638,195) | 37.05(8.23) |
| | SCGA | 97.4(4.80) | $1.34\times10^{-3}(8.72\times10^{-4})$ | 2,518,301(643,129) | 30.27(7.04) |
| | AEGA | 99.4(1.39) | $6.77\times10^{-4}(3.18\times10^{-4})$ | 978,435(71,135) | 10.56(4.81) |
| | Clonal selection algorithm | 90.6(9.95) | $3.15\times10^{-3}(1.47\times10^{-3})$ | 5,075,208(194,376) | 58.02(2.19) |
| | AiNet | 93.8(7.8) | $2.11\times10^{-3}(3.2\times10^{-3})$ | 3,342,864(549,452) | 51.65(6.91) |
| | CAB | **100(2)** | $2.22\times10^{4}(3.1\times10^{3})$ | **680,211(12,547)** | **7.33(1.84)** |
| $f_6$ | Deterministic crowding | 9.37(1.91) | $3.26\times10^{-3}(5.34\times10^{-4})$ | 832,546(75,413) | 4.58(0.57) |
| | Probabilistic crowding | 15.17(2.43) | $2.87\times10^{-3}(5.98\times10^{-4})$ | 1,823,774(265,387) | 12.92(2.01) |
| | Sequential fitness sharing | 15.29(2.14) | $1.42\times10^{-3}(5.29\times10^{-4})$ | 1,767,562(528,317) | 14.12(3.51) |
| | Clearing procedure | **18(0)** | $1.19\times10^{-3}(6.05\times10^{-4})$ | 1,875,729(265,173) | 11.20(2.69) |
| | CBN | 14.84(2.70) | $4.39\times10^{-3}(2.86\times10^{-3})$ | 2,049,225(465,098) | 18.26(4.41) |
| | SCGA | 4.83(0.38) | $1.58\times10^{-3}(4.12\times10^{-4})$ | 2,261,469(315,727) | 13.71(1.84) |
| | AEGA | **18(0)** | $3.34\times10^{-4}(1.27\times10^{-4})$ | 656,639(84,213) | 3.12(1.12) |
| | Clonal selection algorithm | **18(0)** | $3.42\times10^{-3}(1.58\times10^{-3})$ | 4,989,856(618,759) | 33.85(5.36) |
| | AiNet | **18(0)** | $2.11\times10^{-3}(3.31\times10^{-5})$ | 3,012,435(332,561) | 26.32(2.54) |
| | CAB | **18(0)** | $1.02\times10^{4}(4.27\times10^{6})$ | **431,412(21,034)** | **2.21(0.51)** |
| $f_7$ | Deterministic crowding | 52.6(8.86) | $3.71\times10^{-3}(1.54\times10^{-3})$ | 2,386,960(221,982) | 19.10(2.26) |
| | Probabilistic crowding | 79.2(4.94) | $3.48\times10^{-3}(3.79\times10^{-3})$ | 3,861,904(457,862) | 43.53(4.38) |
| | Sequential fitness sharing | 63.0(5.49) | $4.76\times10^{-3}(3.55\times10^{-3})$ | 3,619,057(565,392) | 42.98(6.35) |
| | Clearing procedure | 79.4(4.31) | $2.95\times10^{-3}(1.64\times10^{-3})$ | 3,746,325(594,758) | 45.42(7.64) |
| | CBN | 71.3(9.26) | $3.29\times10^{-3}(4.11\times10^{-3})$ | 4,155,209(465,613) | 48.23(5.42) |
| | SCGA | 94.9(8.18) | $2.63\times10^{-3}(1.81\times10^{-3})$ | 3,629,461(373,382) | 47.84(0.21) |
| | AEGA | 98(2) | $1.31\times10^{-3}(8.76\times10^{-4})$ | 1,723,342(121,043) | 12.54(1.31) |
| | Clonal selection algorithm | 89.2(5.44) | $3.02\times10^{-3}(1.63\times10^{-3})$ | 5,423,739(231,004) | 47.84(6.09) |
| | AiNet | 92.7(3.21) | $2.79\times10^{-3}(3.19\times10^{-4})$ | 4,329,783(167,932) | 41.64(2.65) |
| | CAB | **100(1)** | $3.32\times10^{4}(5.25\times10^{6})$ | **953,832(9,345)** | **8.82(1.51)** |
| $f_8$ | Deterministic crowding | 44.2(7.93) | $4.45\times10^{-3}(3.63\times10^{-3})$ | 2,843,452(353,529) | 23.14(3.85) |
| | Probabilistic crowding | 70.1(8.36) | $2.52\times10^{-3}(1.47\times10^{-3})$ | 4,325,469(574,368) | 49.51(6.72) |
| | Sequential fitness sharing | 58.2(9.48) | $4.14\times10^{-3}(3.31\times10^{-3})$ | 4,416,150(642,415) | 54.43(12.6) |
| | Clearing procedure | 67.5(10.11) | $2.31\times10^{-3}(1.43\times10^{-3})$ | 4,172,462(413,537) | 52.39(7.21) |
| | CBN | 53.1(7.58) | $4.36\times10^{-3}(3.53\times10^{-3})$ | 4,711,925(584,396) | 61.07(8.14) |
| | SCGA | 87.3(9.61) | $3.15\times10^{-3}(2.07\times10^{-3})$ | 3,964,491(432,117) | 53.87(8.46) |
| | AEGA | 90.6(1.65) | $2.55\times10^{-3}(9.55\times10^{-4})$ | 2,213,754(412,538) | 16.21(3.19) |
| | Clonal selection algorithm | 74.4(7.32) | $3.52\times10^{-3}(2.19\times10^{-3})$ | 5,835,452(498,033) | 74.26(5.47) |
| | AiNet | 83.2(6.23) | $3.11\times10^{-3}(2.41\times10^{-4})$ | 4,123,342(213,864) | 60.38(5.21) |
| | CAB | **97(2)** | $1.54\times10^{3}(4.51\times10^{4})$ | **1,121,523(51,732)** | **12.21(2.66)** |

**Table 4.** Performance comparison among multimodal optimization algorithms for the test functions $f_5 - f_8$. The standard unit of the column ET is seconds (numbers in parentheses are standard deviations). Bold-case letters represent best results.

### 4.3. Comparing CAB performance in rough landscapes functions

This section presents the performance comparison among different algorithms solving multimodal optimization problems which are listed in Table 2. Such problems hold lots of local optima and very rugged landscapes. The goal of multimodal optimizers is to find as many as possible global optima and possibly good local optima. Rastrigin's function $f_5$ and Griewank's function $f_7$ have 1 and 18 global optima respectively, becoming practical as to test to whether a multimodal algorithm can find a global optimum and at least 80 higher fitness local optima to validate the algorithms' performance.

Our main objective in these experiments is to determine whether CAB is more efficient and effective than other existing algorithms for finding the multiple high fitness optima of functions $f_5 - f_8$. In the experiments, the initial population size for all algorithms has been set to 1000. For Sequential Fitness Sharing, Clearing Procedure, CBN, Clonal Selection, SCGA, and AEGA, we have set the distance threshold $\sigma_s$ to 5. The algorithms' stopping criterion checks whenever the number of optima found cannot be further increased in 50 successive generations after the first 500 generations. If such condition prevails then the algorithm is halted. We still evaluate the performance of all algorithms using the aforementioned four measures NO, DO, FE, and ET.

Table 4 provides a summary of the performance comparison among different algorithms. From the NO measure, we observe that CAB could always find more optimal solutions for the multimodal problems $f_5 - f_8$. For Rastrigin's function $f_5$, only CAB can find all multiple high fitness optima 49 times out of 50 runs and its average successful rate for each run is higher than 97%. On the contrary, other algorithms cannot find all multiple higher fitness optima for any run. For $f_6$, 5 algorithms (Clearing Procedure, SCGA, AEGA, clonal selection algorithm, AiNet and CAB) can get all multiple higher fitness maxima for each run respectively. For Griewank's function ($f_7$), only CAB can get all multiple higher fitness optima for each run. In case of the modified Griewank's function ($f_8$), it has numerous optima whose value is always the same. However, CAB still can find all global optima with a effectiveness rate of 95% .

From the FE and ET measures in Table 4, we can clearly observe that CAB uses significantly fewer function evaluations and a shorter running time than all other algorithms under the same termination criterion. Moreover, Deterministic Crowding leads to premature convergence as CAB is at least 2.5, 3.8, 4, 3.1, 4.1, 3.7, 1.4, 7.9 and 4.9 times faster than all others respectively according to Table 4 for functions $f_5 - f_8$.

### 5. Conclusions

In recent years, several metaheuristic optimization methods have been inspired from nature-like phenomena. In this article, a new multimodal optimization algorithm known as the Collective Animal Behavior Algorithm (CAB) has been introduced. In CAB, the searcher agents are a group of animals that interact to each other depending on simple behavioral rules which are modeled as mathematical operators. Such operations are applied to each agent considering that the complete group hold a memory to store its own best positions seen so far, using a competition principle.

CAB has been experimentally evaluated over a test suite consisting of 8 benchmark multimodal functions for optimization. The performance of CAB has been compared to some other existing algorithms including Deterministic Crowding [17], Probabilistic Crowding [18], Sequential Fitness Sharing [15], Clearing Procedure [20], Clustering Based Niching (CBN) [19], Species Conserving Genetic Algorithm (SCGA) [21], elitist-population strategies (AEGA) [22], Clonal Selection algorithm [24] and the artificial immune network (aiNet) [25]. All experiments have demonstrated that CAB generally outperforms all other multimodal metaheuristic algorithms regarding efficiency and solution quality, typically showing significant efficiency speedups. The remarkable performance of CAB is due to two different features: (i) operators allow a better  exploration of the search space, increasing the capacity to find multiple optima; (ii) the diversity of solutions contained in the $\mathbf{M}_h$ memory in the context of multimodal optimization, is maintained and even improved through of the use of a competition principle (dominance concept).